\documentclass{article}
\usepackage{ijcai21}

\usepackage{times}

\usepackage{soul}
\usepackage{url}
\usepackage[hidelinks]{hyperref}
\usepackage[utf8]{inputenc}
\usepackage[small]{caption}
\usepackage{graphicx}
\usepackage{amsmath}
\usepackage{booktabs}
\usepackage{multirow}

\title{Semi-supervised Learning for Marked Temporal Point Processes}

\author{
Shivshankar Reddy\and Anand Vir Singh Chauhan\and Maneet Singh \footnote{Contact Author}\and Karamjit Singh\\
\affiliations
AI Garage, Mastercard, India
\emails
\{shivshankar.reddy, anandvirsingh.chauhan, maneet.singh, karamjit.singh\}@mastercard.com
}

\begin{document}

\maketitle

\begin{abstract}
Temporal Point Processes (TPPs) are often used to represent the sequence of events ordered as per the time of occurrence. Owing to their flexible nature, TPPs have been used to model different scenarios and have shown applicability in various real-world applications. While TPPs focus on modeling the event occurrence, \textit{Marked} Temporal Point Process (MTPP) focuses on modeling the category/class of the event as well (termed as the \textit{marker}). Research in MTPP has garnered substantial attention over the past few years, with an extensive focus on supervised algorithms. Despite the research focus, limited attention has been given to the challenging problem of developing solutions in semi-supervised settings, where algorithms have access to a mix of labeled and unlabeled data. This research proposes a novel algorithm for \textit{Semi-supervised Learning for Marked Temporal Point Processes (SSL-MTPP)} applicable in such scenarios. The proposed SSL-MTPP algorithm utilizes a combination of labeled and unlabeled data for learning a robust marker prediction model. The proposed algorithm utilizes an RNN-based Encoder-Decoder module for learning effective representations of the time sequence. The efficacy of the proposed algorithm has been demonstrated via multiple protocols on the Retweet dataset, where the proposed SSL-MTPP demonstrates improved performance in comparison to the traditional supervised learning approach.
\end{abstract}

\section{Introduction}
%Para1: Introduction to TPPs and MTPPs: Highlight the importance, usage and benefits of TPP domain.

A plethora of events happen in a period of twenty four hours which sum up to make a day in the life of a human being. Generally, these events do not occur at continuous time stamps since at different times of the day different activities can be performed. For example, social media usage of an individual involves posts on various platforms at different times during the day. Other examples include visits to the hospitals on different days of the week/month, purchasing groceries/essentials at an e-commerce website, or financial transactions of buying/selling stocks at a stock market exchange. All the above examples have the common characteristic of irregular time between events. Given the presence and need of modeling such data, \textit{Temporal Point Processes (TPPs)} are deemed as an appropriate method to formulate such irregularly time-spaced sequence data, i.e. sequences containing discrete events occurring at irregular times. Beyond modelling the event occurrence time, TPPs also help to model the type of action (or category of the event) also known as the \textit{marker} information. TPPs which model both the event time occurrence and marker information are called \textit{Marked Temporal Point Processes (MTPPs)} \cite{du2016recurrent}. MTPP based models have shown exemplary performance for predicting the next event time and event marker, while demonstrating applicability in several real-world scenarios as well.

\begin{figure}[t]
    \centering
    \includegraphics[width=3in]{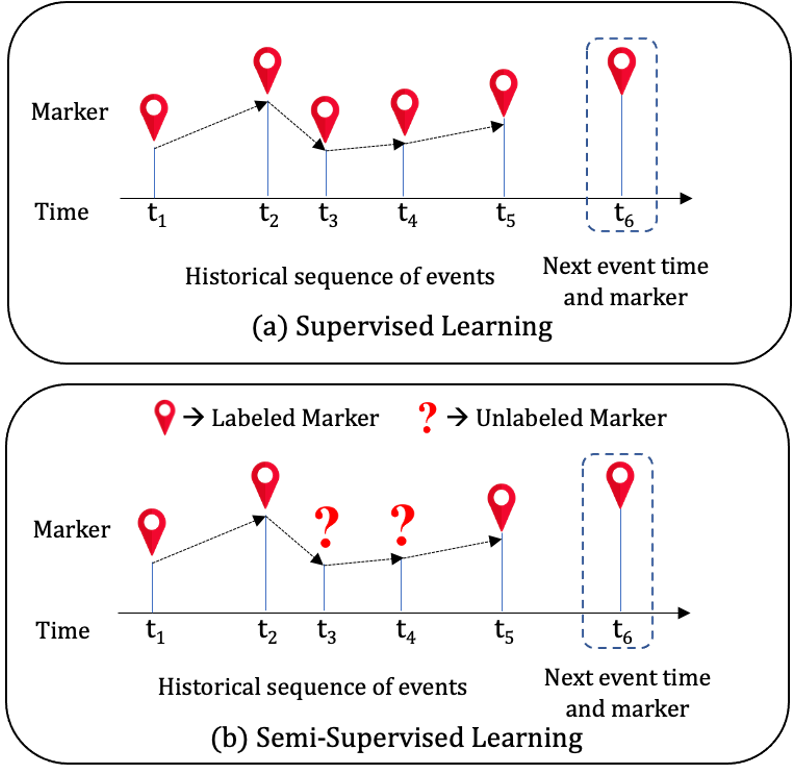}
    \caption{In supervised learning, all the historical events' time and marker data is labeled, whereas in semi-supervised learning, some historical events have unlabeled markers.}
    \label{fig:intro}
\end{figure}

Most of the existing research in the area of MTPPs has focused on utilizing large amount of labeled data for learning a prediction model (Figure \ref{fig:intro}(a)). The dependence on such training sets often makes it challenging to utilize the models for real-world setups having limited availability of labeled training data. Recently, due to the technological advancements and heavy reliance on the digital world, large amount of data is generated every day; a majority of which is unlabeled in nature. For example, patient visits to the hospital can be recorded instantly, however, the severity of the disease or duration of stay (marker) will be known at a later time. In order to benefit from the abundant data, the concept of semi-supervised learning \cite{zhu2005semi} has been adopted for learning prediction models, which is the branch of machine learning utilizing both labeled and unlabeled data with the goal of increasing the performance even under limited labeled training data \cite{van2020survey}.

Despite the advances in the field of semi-supervised learning and marked temporal point processes independently, to the best of our knowledge, no research has focused at the intersection of the two fields. Owing to the limitation of limited labeled training data availability in real-world setups, it is important to develop new techniques incorporating the semi-supervised learning approach into MTPP based models (Figure \ref{fig:intro}(b)). In the context of MTPPs, the concept of supervised learning is incorporated with regard to the labeled marker information and not the time of the event occurrence. To this effect, this research proposes a novel algorithm for \textit{Semi-supervised Learning for Marked Temporal Point Processes (SSL-MTPP)} which utilizes a combination of labeled and unlabeled data for learning a robust event marker prediction model. The proposed SSL-MTPP algorithm has been evaluated on six protocols containing varying labeled data ranging from $10K$ to $0.7M$ events, where it demonstrates improved performance over the baseline supervised learning approach. As part of this research, analysis is also performed on the Macro-F1 and Micro-F1 metrics, while suggesting that Average Precision could be a better metric for capturing the performance for highly imbalanced per-class distribution datasets.

%The remainder of this paper has been structured in the following manner: Section \ref{sec:related} presents the related work in the field of Temporal Point Processes and Semi-supervised Learning, Section \ref{sec:proposed} presents the proposed SSL-MTPP algorithm along with the implementation details. Section \ref{sec:experiment} presents the details regarding the experiments and analysis, while Section \ref{sec:conclusion} concludes the paper with the conclusions and discussions regarding the future direction of this research.

\section{Related Work}
\label{sec:related}

This research presents a novel SSL-MTPP algorithm which operates at the intersection of (i) Marked Temporal Point Processes and (ii) Semi-supervised Learning. To the best of our knowledge, this is the first research focusing on this challenging area. Due to the lack of specific literature on the same area of focus, the following subsections thus independently elaborate upon these two areas of research.

\subsection{Marked Temporal Point Process (MTPP)}

Initially, research in the field of Temporal Point Processes (TPP) focused primarily on Self-Exciting \cite{hawkes1971spectra} and Self-Correcting \cite{isham1979self} processes. While such techniques demonstrated good performance, they were deemed to have limited flexibility limiting their usage in real-world scenarios. A significant breakthrough in the TPP domain happened in the last decade with the advent of deep learning by the usage of neural-networks for modelling the underlying conditional function. Traditionally, TPPs were used to only model the time occurrence of an event, however, later they were also used to model and predict the corresponding type/class of the event called the \textit{marker}. Such techniques which focused on predicting the event and corresponding type were termed as modeling \textit{Marked Temporal Point Processes (MTPP)} \cite{du2016recurrent}. Recently, Recurrent Neural Networks (RNN) have also been used to model the Marked Temporal Point Processes. For example, techniques such as the Recurrent Marked Temporal Point Processes \cite{du2016recurrent} and Neural Hawkes Process \cite{mei2016neural} utilize the RNN module as the backbone architecture. Similarly, a fully neural network based model has also been proposed for general TPPs \cite{omi2019fully} to model the cumulative intensity function with a neural network. Further, recently, techniques have also been proposed to learn robust representations under the constraint of missing observations as well \cite{gupta2021learning}. Most of the techniques mentioned above follow the supervised learning approach which often require large amount of labeled training data.
%Add line that all of this is supervised in nature requiring large amount of data.

%but the problem with these approaches was that they were not flexible. The real breakthrough in Temporal Point Processes happened in last decade with the advent of neural-network based Temporal Point Processes models which incorporated the conditional intensity function in the modelling pipeline. Traditionally, Temporal Point Processes were used to only model the time occurrence of an event but later they were also used to model and predict the corresponding type/class of the event called the marker and thus Marked Temporal Point Processes came into the picture. In recent times, there also has been a trend to use Recurrent Neural Networks to model the Marked Temporal Point Processes by Recurrent Marked Temporal Point Processes \cite{du2016recurrent}, Neural Hawkes Process \cite{mei2016neural}. Recurrent Marked Temporal Point Processes is also the baseline research work for the experiments discussed later. Along similar lines of using a RNN to model the conditional intensity function, Fully Neural Network based Model for General Temporal Point Processes \cite{omi2019fully} also developed a novel RNN model called FullyNN to model the cumulative intensity function with a neural net. 

\begin{figure}[t]
    \centering 
    \includegraphics[width=3.2in]{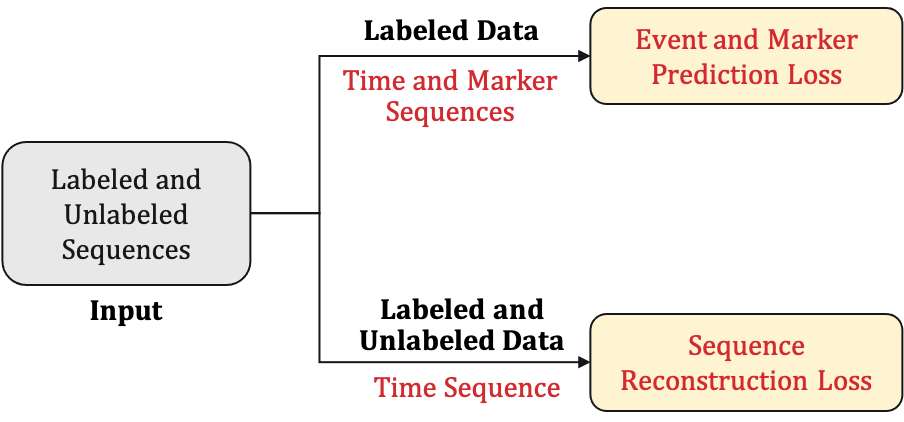}
    \caption{Broad overview of the proposed semi-supervised learning algorithm for marked temporal point processes. The model is trained with a combination of marker/event prediction loss (supervised) and event sequence reconstruction loss (unsupervised).}
    \label{fig:overview}
\end{figure}

\subsection{Semi-Supervised Learning}

Broadly, machine learning algorithms can be categorized into the following three categories: (i) supervised learning, (ii) unsupervised learning, and (iii) semi-supervised learning \cite{zhu2005semi}. Semi-supervised learning is the branch of machine learning which attempts to utilize both labeled and unlabeled data during training. Often, the goal is to enhance the performance by incorporating the benefits of both kinds of data. Over the past decades, various algorithms for semi-supervised learning have been proposed. They are broadly classified into two categories: (i) inductive learning and (ii) transductive learning \cite{chapelle2009semi}. In inductive learning based semi-supervised approaches, the goal is to infer the correct mapping from $X$ (data points which are labeled) to $Y$ (corresponding labels) in the form of a classifier, whereas in transductive learning based semi-supervised techniques the goal is to infer the correct labels for the unlabeled data points only. 
%Also, inductive learning not only produces the labels for unlabeled data points but regenerate a classifier as well whereas transductive learning produces labels for unlabeled data points only but it does not generate a classifier. A good analogy from \cite{zhu2005semi} define inductive learning as an in-class exam and transductive learning as take-home exam. %Should I elaborate this example more given that citation is already mentioned?
Inductive learning is further divided into three categories namely unsupervised processing, wrapper methods, and intrinsically semi-supervised learning whereas transductive learning is mostly used in graph-based algorithms like graph construction, graph weighting and graph inference \cite{van2020survey}.

%One paragraph to combine the above paragraphs
The proposed SSL-MTPP model extends the current body of literature by focusing on the intersection of the above two fields of research. Specifically, current TPP models follow supervised learning approach which often require large amount of labeled data. In real world scenarios, such data is often not easily available. In order to overcome the limitation of limited labeled data availability, this research proposes a novel SSL-MTPP algorithm which follows the inductive approach of semi-supervised learning and incorporates labeled and unlabeled data for learning an marker prediction model.

%Traditionally, machine learning algorithms followed two methods: (i) Supervised learning: It is the branch of machine learning in which the data is labeled i.e. the input data points have a corresponding output label and the task is to use a classification or regression method to predict the next output label. (ii) Unsupervised Learning: In unsupervised learning, the data is not labeled i.e the input data points are not provided with label of the output and the machine learning algorithms use the input data points to find a similar pattern in them and then cluster them in separate groups. Since, large amount of labeled data is not easily available, the concept of Semi-Supervised learning was evolved \cite{zhu2005semi}. 

\begin{figure*}[t]
    \centering 
    \includegraphics[width=6.5in]{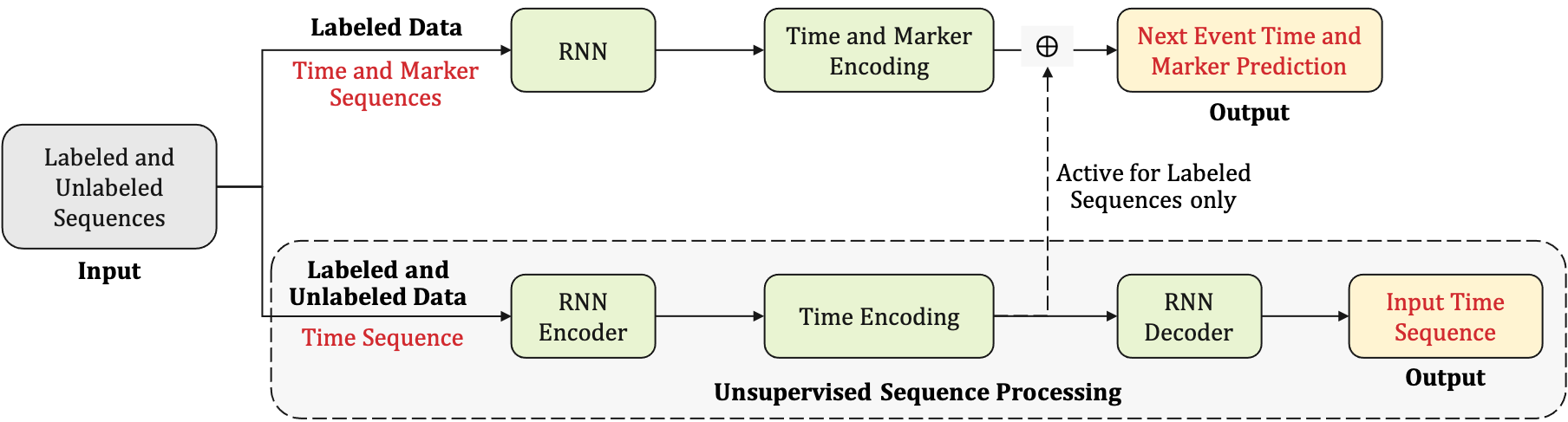}
    \caption{Diagrammatic representation of the proposed semi-supervised learning algorithm for marked temporal point processes. Given a pool of labeled and unlabeled event sequences, the proposed algorithm focuses on learning meaningful time and marker encodings, useful for the marker prediction. The unlabeled sequences are passed through the unsupervised sequence processing path utilizing an RNN encoder-decoder module. The labeled sequences are used for learning the inter-dependence between the marker and the event occurrences, followed by the next marker prediction. }
    \label{fig:algo}
\end{figure*}

\section{Proposed Algorithm}
\label{sec:proposed}

This research proposes a novel algorithm for \textit{Semi-supervised Learning for Marked Temporal Point Processes (SSL-MTPP)}. The proposed SSL-MTPP algorithm is useful in scenarios which contain additional unlabeled time sequences which can be used for improved model learning for marker prediction. Here, the notion of supervision is with respect to the marker information (class label) and not with respect to the event occurrence. Figure \ref{fig:overview} presents an overview of the proposed SSL-MTPP algorithm. The model is trained via a combination of event/marker prediction loss terms and a reconstruction loss on the event sequences. The top path (labeled) allows the model to learn the relationship between the markers and events, while the bottom path (unlabeled) enables learning of a meaningful embedding for the given input event sequence which is fused with the embedding obtained from the labeled data for a more robust representation. Mathematically, the loss of the SSL-MTPP algorithm (Figure \ref{fig:overview}) is given as:
\begin{equation}
    \mathcal{L}_{SSL-MTPP} = \mathcal{L}_{Time} + \mathcal{L}_{Marker} + \mathcal{L}_{Recon.}
\end{equation}
where, $\mathcal{L}_{Time}$ and $\mathcal{L}_{Marker}$ refer to the terms corresponding to the next event time and marker prediction, respectively, while the $\mathcal{L}_{Recon.}$ term refers to the reconstruction component. As shown in Figure \ref{fig:overview}, the next event time and marker prediction requires labeled data (i.e. the $\mathcal{L}_{Marker}$ and $\mathcal{L}_{Time}$ loss components), while the reconstruction loss ($\mathcal{L}_{Recon.}$) is unsupervised with respect to the marker information, and thus does not utilize the marker information during training. 
%With respect to the task of marker prediction, in the proposed architecture, the $\mathcal{L}_{Marker}$ component requires labeled data corresponding to the previous markers and the next marker, while the $\mathcal{L}_{Time}$ and $\mathcal{L}_{Recon.}$ terms are unsupervised in nature since they do not require the label information corresponding to the markers. 
Detailed description of the entire architecture and each component of the proposed SSL-MTPP algorithm is as below.

\subsection{SSL-MTPP Algorithm}

Figure \ref{fig:algo} presents the architecture of the proposed algorithm consisting of a labeled and an unlabeled path. The labeled path takes as input of a given set of sequences ($\mathcal{S}$) consisting of $n$ sequence pairs ($x_i, y_i$) having the event time information ($x_i$) and the marker information ($y_i$). It utilizes a Recurrent Neural Network (RNN) architecture for learning an embedding capturing the behavior of the marker and the time occurrence in the given sequences. The embedding is used by two independent modules for (i) marker prediction and (ii) time prediction of the next event. On the other hand, the unlabeled path utilizes an RNN Encoder-Decoder model \cite{cho2014learning} for learning an embedding for the time sequence only. The learned embedding represents the time sequence and is used to supplement the marker-time embedding for a robust representation. Mathematical formulation and explanation of the different steps and models involved along with each loss component is as follows:  \\

\noindent \textbf{Unsupervised Reconstruction Loss ($\mathcal{L}_{Recon.}$) Component:} The reconstruction loss is applied on the event sequences in the training set, without utilizing the marker information. Therefore, as shown in Figure \ref{fig:algo}, the reconstruction loss is applied on the unlabeled samples and the labeled samples (without utilizing the marker). Given $n$ sequences in the training set $\mathcal{S} = \{x_1, x_2, ... x_n\}$, where each sequence $x_i$ contains time occurrence of $k$ events, the reconstruction loss is as:
\begin{equation}
    \mathcal{L}_{Recon.} = \sum_{i=1}^n\|x_i - \mathcal{D}(\mathcal{E}(x_i))\|_2^2
    \label{eq:recon}
\end{equation}
where, $\mathcal{E}(.)$ and $\mathcal{D}(.)$ correspond to the encoder and decoder RNN modules. The reconstruction loss focuses on learning a meaningful representation for the given time sequence, which is used for subsequent marker prediction. The reconstruction module is unsupervised in nature and does not utilize the class labels (markers) during training. A robust encoding of the time sequence ($\mathcal{E}(x_i)$) is used to augment the learned representation of the SSL-MTPP model for enhanced marker prediction. Details regarding the utilization of the encoder embedding ($\mathcal{E}(x_i)$) for marker prediction are given below: \\

\noindent \textbf{Supervised Marker ($\mathcal{L}_{Marker}$) and Time ($\mathcal{L}_{Time}$) Prediction Loss Components:} An input sequence pair ($x_i, y_i$) containing the event time information and the marker information, respectively, is passed to an RNN module for obtaining a feature representation modeling the inter-dependence of the marker and time information:
\begin{equation}
    f_i = RNN(x_i, y_i)
\end{equation}
The extracted representation ($f_i$) is then combined with the representation obtained earlier from the encoder module ($\mathcal{E}(x_i)$) to generate a fused embedding as follows:
\begin{equation}
    f^{fused}_i = f_i + \lambda* \mathcal{E}(x_i)
    \label{eq:sum}
\end{equation}
where, $\lambda$ corresponds to the weight given to the representation obtained via the RNN encoder. By fusing the two representations, the proposed algorithm is able to benefit from modeling the inter-dependence between the marker and time sequence ($f_i$) in a supervised manner and learning a robust representation of only the time sequence in an unsupervised manner ($\mathcal{E}(x_i)$). The fused representation ($f^{fused}_i$) is then provided to two multi-layer perceptron networks for predicting the next event time and marker category. The prediction models are trained with the following loss components for each event in the training set:
\begin{equation}
    \mathcal{L}_{Marker} = - \sum_{c=1}^{M} y_{i,c}^{j}log(p_{i,c}^{j}); \mathcal{L}_{Time} = \|x_i^j - x^{j}{'}_{i}\|
\end{equation}
where, $\mathcal{L}_{Marker}$ corresponds to the cross-entropy loss for multi-class marker classification, while $\mathcal{L}_{Time}$ corresponds to the mean absolute error loss for predicting the next event time occurrence as a regression task. Here, prediction is performed for an event $j$ belonging to sequence $i$ (i.e. event time $x_i^j$ and marker $y_i^j$) for a $M$ class prediction setup. $y_{i,c}^j$ is a binary variable signifying whether sample $y_i^j$ is of class $c$ or not, while $p_{i,c}^j$ is the probability of the sample belonging to class $c$, and $x^{j}{'}_i$ is the predicted event time occurrence for the given event.

\subsection{Implementation Details}

The proposed SSL-MTPP model has been implemented in the PyTorch environment \cite{pytorch}. SSL-MTPP algorithm utilizes the RMTPP architecture \cite{du2016recurrent} as the base model. The supervised branch contains a five layer Long Short-Term Memory (LSTM) module \cite{hochreiter1997long}, while the unsupervised branch contains a two layer RNN encoder and decoder. The marker and event prediction modules contain two dense layers, respectively. Dropout \cite{dropout} has also been applied as a regularizer after the RNN layer. The weight value ($\lambda$ in Equation \ref{eq:sum}) has been set to $0.1$. The model has been trained for 100 epochs with a learning rate of 0.01 using the Adam optimizer \cite{adam}. Training has been performed with a batch-size of $1024$ sequences on an NVIDIA Quadro RTX6000 GPU.

\begin{table}[]
    \centering
    \begin{tabular}{|c|c|c|}
         \hline
         \textbf{Protocol} & \textbf{Labeled Data} & \textbf{Unlabeled Data} \\
         \hline
         \hline
         P-1 & 10K & 1.39M \\
         \hline
         P-2 & 20K & 1.38M \\
         \hline
         P-3 & 30K & 1.37M \\
         \hline
         P-4 & 50K & 1.35M \\
         \hline
         P-5 & 140K (10\%) & 1.26M \\
         \hline
         P-6 & 0.7M (50\%) & 0.7M \\
         \hline
    \end{tabular}
    \caption{Six protocols have been used to evaluate the proposed SSL-MTPP algorithm with varying labeled data (from 10K to 0.7M) and varying unlabeled data on the given training set.}
    \label{tab:protocol}
\end{table}

\section{Experiments and Analysis}
\label{sec:experiment}
The proposed SSL-MTPP algorithm has been evaluated under varying amount of labeled training data. Comparison has been performed with the baseline/native supervised MTPP model which follows a similar architecture as the proposed SSL-MTPP model without the unsupervised branch (described above). Details regarding the dataset, protocols, and results are given below.

\subsection{Dataset and Protocol}

Experiments have been performed on the Retweet dataset \cite{zhao2015seismic}, which is formed through the Seismic dataset. The dataset contains multiple sequences of retweets, where each sequence corresponds to information regarding the retweets on a particular tweet. Each sequence contains information regarding the event time (retweet) and the corresponding marker information. Here, the marker refers to type of user (based on the number of followers) who has retweeted. Three categories of marker are provided: (i) normal user, (ii) influencer user, and (iii) celebrity user. The marker information is defined based on the number of followers (degree) of a given user: (i) degree lower than the median (normal user), (ii) degree higher or equal to the median but less than the $95^{th}$ percentile (influencer user), and (iii) degree higher or equal to the $95^{th}$ percentile (celebrity user). The dataset consists of over two million events with an average sequence length of 209 events.The dataset has imbalance class distribution with 50.6\% event marker are normal users,45\% events marker are influencer users,while only 4.4\% events marker are celebrity users. For experiments, data pertaining to 1.4M events was used for training, while the remaining 60K events formed the test set.

Table \ref{tab:protocol} presents the six protocols used to evaluate the proposed SSL-MTPP algorithm. The training set (consisting of 1.4M events) is split into a labeled set and an unlabeled set for each protocol. The labeled set contains data varying from 10K (P-1) to 0.7M (P-6), while the remaining data forms the unlabeled set. For experiments, the proposed SSL-MTPP algorithm is trained for each protocol, and comparison has been performed with the baseline supervised MTPP model trained in the traditional supervised manner with labeled data only.

\begin{table*}[!t]
    \centering
    \caption{Performance of the proposed semi-supervised learning algorithm with varying amount of labeled data during training.}
    \begin{tabular}{|c|c||l|c|c|c|}
    \hline
    \textbf{Protocol} & \textbf{Labeled Data} & \textbf{Model} & \textbf{Avg. Precision (\%)} & \textbf{Macro-F1 (\%)} & \textbf{Micro-F1 (\%)} \\
    \hline
    \hline
    \multirow{2}{*}{P-1} & \multirow{2}{*}{10K} & Native Supervised MTPP & 38.84 & 39.40 & 58.98 \\
    \cline{3-6}
     & & Proposed Semi-Supervised MTPP& 39.10 & 39.48 & 59.40\\
     \hline
     \hline
    \multirow{2}{*}{P-2} & \multirow{2}{*}{20K} & Native Supervised MTPP & 38.77 & 39.52 & 58.78 \\
    \cline{3-6}
     & & Proposed Semi-Supervised MTPP& 67.95 & 40.77 & 59.50 \\
     \hline
     \hline
    \multirow{2}{*}{P-3} & \multirow{2}{*}{30K} & Native Supervised MTPP & 43.92 & 40.26 & 58.03 \\
    \cline{3-6}
     & & Proposed Semi-Supervised MTPP& 68.07 & 40.72 & 59.56\\
     \hline
     \hline
    \multirow{2}{*}{P-4} & \multirow{2}{*}{50K} & Native Supervised MTPP & 44.91 & 40.74 & 57.61 \\
    \cline{3-6}
     & & Proposed Semi-Supervised MTPP& 68.35 & 40.71 & 59.59 \\
     \hline
     \hline
    \multirow{2}{*}{P-5} & \multirow{2}{*}{140K (10\%)} & Native Supervised MTPP & 45.08 & 37.88 & 56.86 \\
    \cline{3-6}
     & & Proposed Semi-Supervised MTPP & 68.42 & 40.73 & 59.79 \\
     \hline
     \hline

    \multirow{2}{*}{P-6} & \multirow{2}{*}{0.7M (50\%)} & Native Supervised MTPP & 66.74 & 40.47 & 57.76\\
    \cline{3-6}
     & & Proposed Semi-Supervised MTPP& 69.49 & 40.09 & 59.22 \\
     \hline

    \end{tabular}
    \label{tab:res}
\end{table*}

\subsection{Results and Analysis}

Table \ref{tab:res} presents the results obtained on the Retweet dataset. The proposed SSL-MTPP algorithm has been evaluated on different protocols containing varying amount of labeled data (Table \ref{tab:protocol}). In the literature, most of the research has focused on reporting the Macro-F1 and Micro-F1 performance metrics. As part of this research, we observe that these might not be the most appropriate metrics to judge a model's performance, especially under the scenario of imbalanced per-class data. To this effect, along with the Macro-F1 and Micro-F1 values, we also report the average precision of each model. 

As can be observed from Table \ref{tab:res}, limited variation is observed for the Macro-F1 and Micro-F1 values for the proposed and native supervised MTPP model. For the SSL-MTPP model, the Micro-F1 values lie in the range of $59.22\% - 59.59\%$, regardless of the amount of labeled data, and the Macro-F1 values lie in the range of $39.48\% - 40.77\%$, thus demonstrating limited variation. On the other hand, the average precision lies in the range of $39.10\% - 69.49\%$ with varying labeled data. Similar behavior is observed for the native supervised MTPP model as well, where limited variations are observed for the Macro-F1 and Micro-F1 values, while higher range is observed for the average precision metric. The consistent behavior across the two models suggest average precision to be a better metric for comparing performance in the setup of imbalanced testing data across classes. 

\begin{figure}[t]
    \centering 
    \includegraphics[width=3.3in]{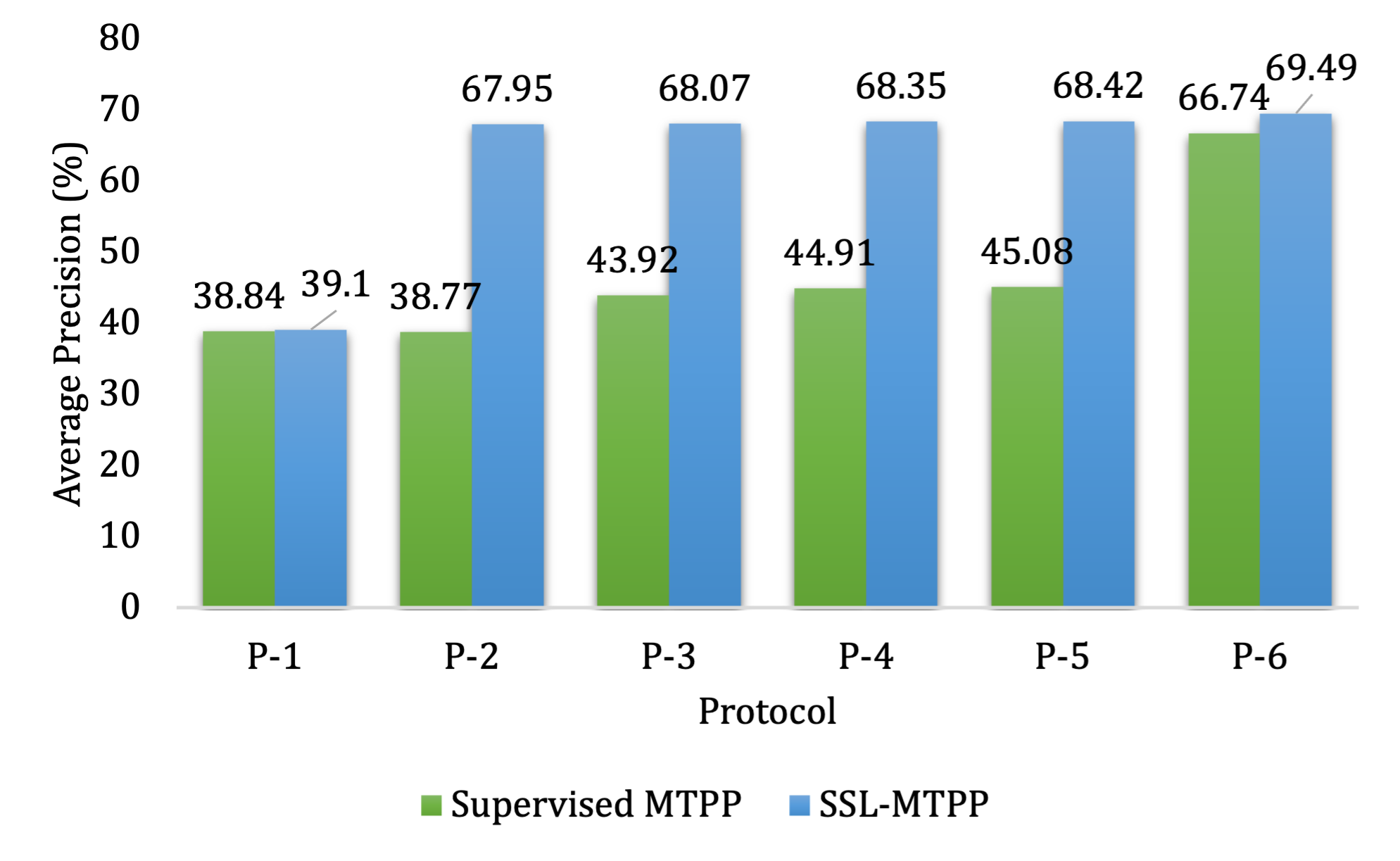}
    \caption{Average precision of the proposed SSL-MTPP model and the native supervised MTPP model on different protocols. The proposed algorithm presents improved performance across protocols. }
    \label{fig:prec}
\end{figure}

Figure \ref{fig:prec} presents the average precision (\%) of the proposed SSL-MTPP model and the native supervised MTPP model for six protocols. Across the protocols, the proposed SSL-MTPP model demonstrates improved performance as compared to the baseline model. In P-1, where only 10K labeled data is available, the SSL-MTPP model obtains an average precision of 39.10\%, presenting an improvement over the baseline model (38.84\%). Larger improvements are observed for P-2 to P-5, where at least 20K labeled data is available for training. For example, in P-4, the SSL-MTPP model obtains an average precision of 68.35\% demonstrating an improvement of almost 24\% as compared to the baseline model. Improvement is also observed for the Macro-F1 and Micro-F1 values across protocols. Further, relatively less improvement is seen when 0.7M labeled data is available for training (P-6), where the proposed SSL-MTPP model achieves an average precision of 69.49\% (as compared to 66.74\% of the baseline model), thus suggesting higher improvement when limited training data is available for training. The above behavior appears intuitive in nature as well, since as the amount of labeled data increases, the baseline model is able to learn better, thus reducing the gap in improvement.

\begin{figure}[t]
    \centering 
    \includegraphics[width=3.3in]{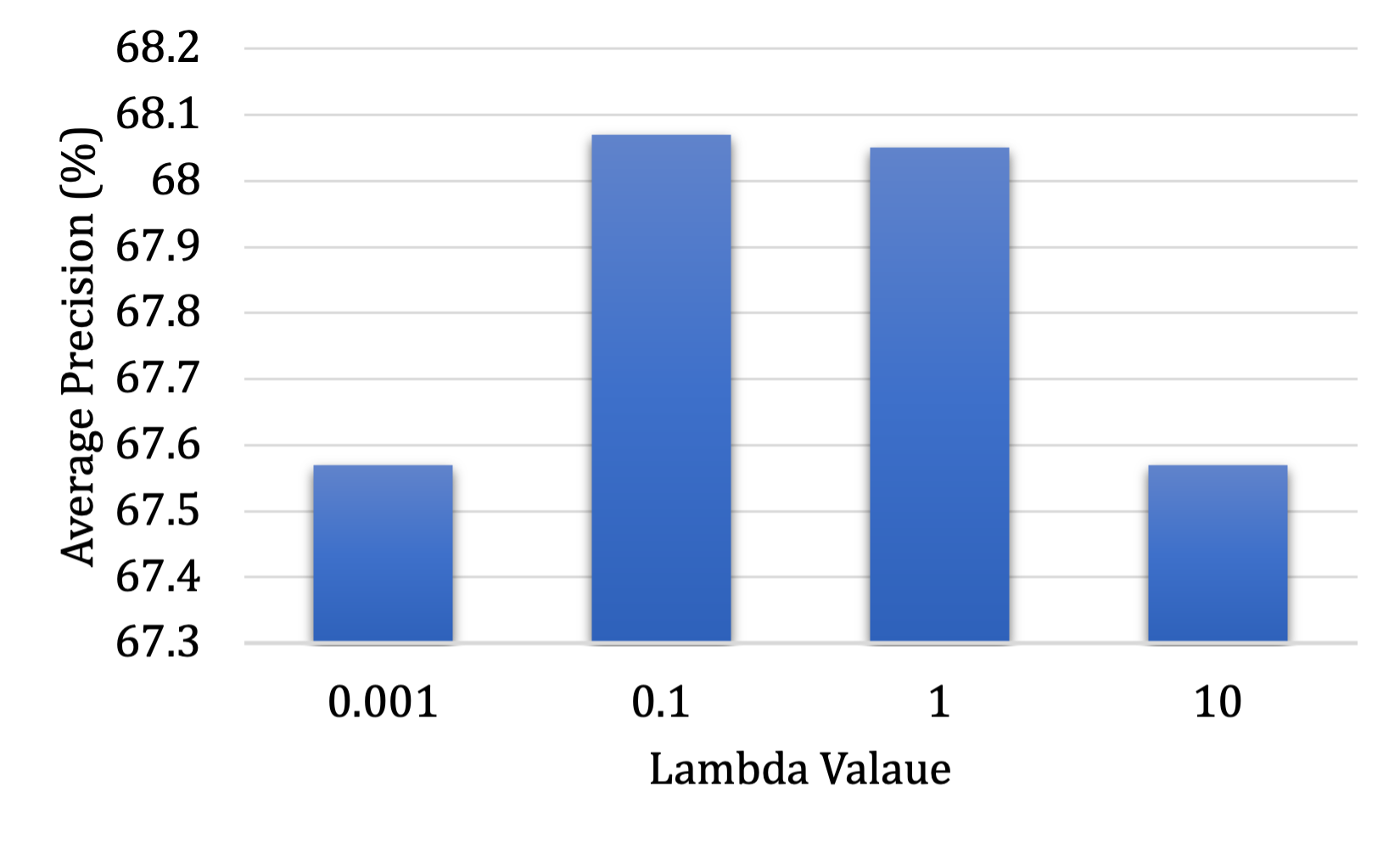}
    \caption{Effect of varying lambda values ($\lambda$ in Equation \ref{eq:sum}) on the average precision performance. Experiments have been performed with the SSL-MTPP algorithm for P-3 (30K labeled training data).}
    \label{fig:bar}
\end{figure}

Experiments have also been performed to analyze the effect of the weight parameter ($\lambda$ in Equation \ref{eq:sum}) for understanding the effect of the fusion of the supervised and unsupervised representations. Figure \ref{fig:bar} presents the average precision obtained on P-3 (30K labeled training data) with varying $\lambda$ values. Best performance is obtained with a value of $0.1$, while a drop in performance is seen with a very small ($0.001$) and a very large ($10$) value. A small value reduces the contribution of the unsupervised representation, while a very large value offsets the contribution of the supervised embedding thus resulting in a drop in performance.

\section{Conclusion and Future Work}
\label{sec:conclusion}

This research proposes a novel semi-supervised learning approach for the marked temporal point processes. The proposed SSL-MTPP algorithm is useful in scenarios where limited labeled data is available for training a marked temporal point process model. In the proposed SSL-MTPP algorithm, the labeled data is passed through an RNN module to generate meaningful embedding which captures the inter-dependence between the time and marker information, and the unlabeled data is passed through an RNN based encoder-decoder to learn the embedding from the time sequence. The representations from both the networks are combined and then used for the next event marker and time prediction. Experiments have been performed on the Retweet dataset on six different protocols by varying the amount of labeled training data. Improved performance is obtained as compared to the baseline supervised learning technique, thus suggesting applicability in different real-world scenarios. Further, analysis is also performed to suggest that the Average Precision metric is more reliable than the Macro-F1 and Micro-F1 metrics in scenarios of imbalanced per-class data.

As part of the future work, the following three directions of research have been identified. Firstly, as observed from Figure \ref{fig:prec}, when only 10K labeled data is available (P-1), the improvement in performance is less than 1\% over the supervised MTPP model. The proposed SSL-MTPP can further be improved to enhance the model performance even when minimal labeled data is available. Secondly, the current architecture assumes that only the marker information to be unlabeled. As a future step, a generic framework can be developed which can handle unlabeled data either in the time or marker information in the historical sequence. Finally, the proposed SSL-MTPP algorithm can be extended to other applications in the TPP domain as well.

\bibliographystyle{named}
\bibliography{bibFile}

\end{document}